\documentclass{article}
\usepackage{spconf,amsmath,graphicx,amsfonts,booktabs,multirow,color}
\usepackage[T1]{fontenc}

\title{mining false positive examples for text-based person re-identification}
\name{Wenhao Xu, Zhiyin Shao, Changxing Ding}
\address{South China University of Technology\\
	School of Electronic and Information Engineering\\
	Guangzhou, Guangdong, China}
\begin{document}
%
\maketitle
\begin{abstract}
Text-based person re-identification (ReID) aims to identify images of the targeted person from a large-scale person image database according to a given textual description. However, due to significant inter-modal gaps, text-based person ReID remains a challenging problem. Most existing methods generally rely heavily on the similarity contributed by matched word-region pairs, while neglecting mismatched word-region pairs which may play a decisive role. Accordingly, we propose to mine false positive examples (MFPE) via a jointly optimized multi-branch architecture to handle this problem. MFPE contains three branches including a false positive mining (FPM) branch to highlight the role of mismatched word-region pairs. Besides, MFPE delicately designs a cross-relu loss to increase the gap of similarity scores between matched and mismatched word-region pairs. Extensive experiments on CUHK-PEDES demonstrate the superior effectiveness of MFPE. Our code is released at {\color{cyan} https://github.com/xx-adeline/MFPE}.
\end{abstract}
\section{introduce}
\label{sec:introduce}
Text-based person re-identification (ReID) is aimed at identifying the targeted person images from a large-scale person image database according to a given textual description. It is a powerful video surveillance tool and has drawn increasing attention from both academia and industry recently. 

\begin{figure}[t]
	\begin{minipage}[t]{1.0\linewidth}
		\centering
		\centerline{\includegraphics[width=8.5cm]{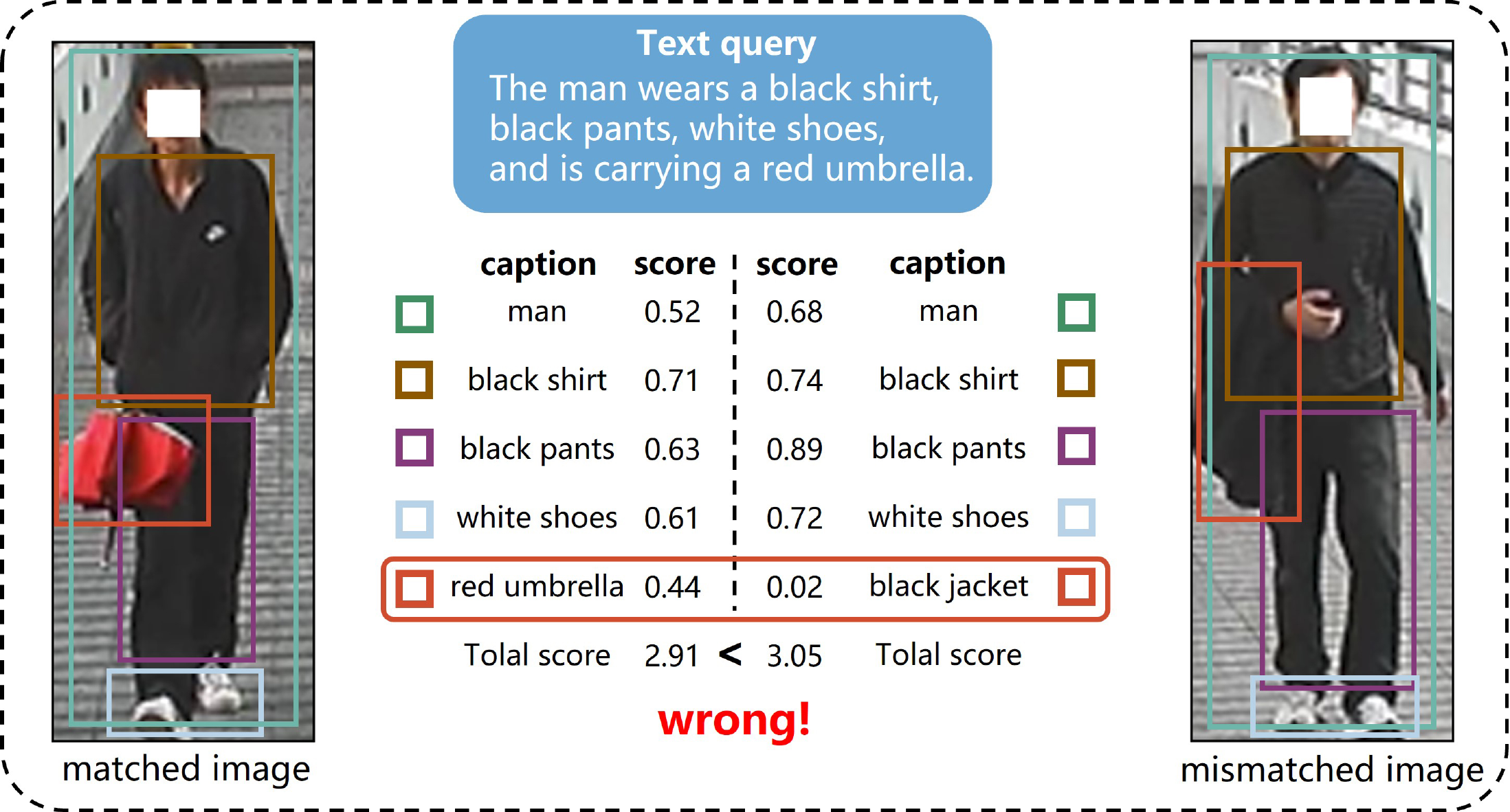}}
	\end{minipage}
	\caption{The two people are difficult to distinguish based on the text query. The person on the left has the same identity as the text, while the person on the right has a different identity.}
	\label{intro}
	\vspace{-1.0em}
\end{figure}

Unfortunately, text-based person ReID is still a challenging problem due to fine-grained problem. To be specific, it is difficult to distinguish between people who are dressed very similarly, such as the two people shown in Fig. \ref{intro}. They wear black shirts, black pants, and white shoes as described in the text query. The only difference is that the man on the left is holding a red umbrella and the man on the right is holding a jacket. In this case, most existing methods generally ignore a problem \cite{ding2021semantically}, \cite{MIA}, \cite{CMAMM}, \cite{zhu2021dssl}, \cite{NAFS}, \cite{HGAN}, \cite{TDE}. The matched word-region pairs get high similarity scores and contribute a lot to the final instance-level similarity, while the mismatched word-region pairs have little effect on it. This means that all mismatched word-region pairs are probably to be completely ignored. However, a matched image-text pair should not have any mismatched word-region pairs. The ignored word-region pairs are inevitably prone to cause false-positive matching. Therefore, it is essential to highlight the role of mismatched word-region pairs to downgrade the overall similarity of mismatched image-text pairs.

To this end, we propose to mine false positive examples (MFPE) via a jointly optimized multi-branch architecture for text-based person ReID. As illustrated in Fig. \ref{method}, MFPE contains three branches. The global and local branches are for aligning visual features and text features and calculating the similarity of image-text pairs \cite{ding2021semantically}. In the false positive mining (FPM) branch, we delicately design a novel cross-relu loss to forcedly delineate a clear decision boundary and maximally increase the gap of similarity scores between matched and mismatched word-region pairs which are sampled in a balanced manner. Based on the discriminative similarity, we can precisely mine mismatched word-region pairs and utilize them as a bias to modify the overall similarity. To demonstrate the efficacy of MFPE, we conduct extensive experiments on the CUHK-PEDES database. The results show that MFPE can effectively mine false positive examples and outperforms compared methods.
\section{related works}
\label{sec:related}
Text-based person ReID was first introduced by Li et al. \cite{Li_2017_CVPR}. They proposed a GNA-RNN model to calculate the affinity between each image-text pair and collected a large-scale person description dataset called CUHK-PEDES. Later, to alleviate the fine-grained problem caused by all samples belonging to a single category, the region-word-based method and the region-phrase-based method are popular for text-based person ReID. For example, Niu et al. \cite{MIA} proposed a Multi-granularity Image-text Alignments (MIA) model to align cross-modal features at three different granularities. Wang et al. \cite{wang2020vitaa} aligned body regions with noun phrases with the help of a light auxiliary attribute segmentation layer and a natural language parser. SSAN \cite{ding2021semantically} was proposed to automatically extract region-level textual features for its corresponding visual regions by introducing extra prediction to the word-region correspondences. Zhu et al. \cite{zhu2021dssl} proposed a DSSL model to explicitly separate surroundings information and person information to obtain higher retrieval accuracy. Recently, Wang et al.

Recently, Zhang et al. \cite{zhang2022negative} proposed a negative-aware attention framework. However, we found through experiments that as a model for image-text retrieval, its data sampling method and attention mechanism are not suitable for text-based ReID. For example, NAAF judges whether the word-region matches based on the principle that all word-region pairs of a mismatched image-text pair mismatch. It is obviously inappropriate for person REID because all samples belong to the same category, and regions generally have matched words. Inspired by this, we explore mining false positive examples for text-based ReID.
\section{Method}
\label{sec:method}
\subsection{Global and local branches}
{\bf Visual Representation Extraction:} We utilize a pretrained ResNet-50 \cite{he2016deep} to extract visual feature maps $F\in \mathbb{R} ^{H\times W\times C} $. To obtain the global visual representation $V_{g} \in \mathbb{R} ^{P} $, we first perform Global Max Pooling (GMP) to downsample on $F$. And then we reduce it to $P$-dim through a 1$\times$1 conv layer. For the local branch \cite{ding2021semantically}, we first horizontally partition $F$ into $K$ non-overlapping regions $V_{e}= \left \{v_{k} \right \}_{k=1}^{K}$ \cite{PCB} and separately embed them through $K$ corresponding 1$\times$1 conv layers.

\noindent{\bf Textual Representation Extraction:} We utilize a Bi-LSTM \cite{lstm} to extract the sentence representation $E \in \mathbb{R} ^{C\times n} $ after the words are embedded via a pretrained BERT language model \cite{bert}. Similar to the global visual branch, we perform Row-wise Max Pooling (RMP) and a 1$\times$1 conv layer to obtain the global textual representation $T_{g}  \in \mathbb{R} ^{P} $. For the local branch, we adopt a Word Attention Module (WAM) \cite{ding2021semantically} to modify $E$ to $K$ local textual representations.\begin{figure*}[t]
	\begin{minipage}[t]{1.0\linewidth}
		\centering
		\centerline{\includegraphics[width=17.5cm]{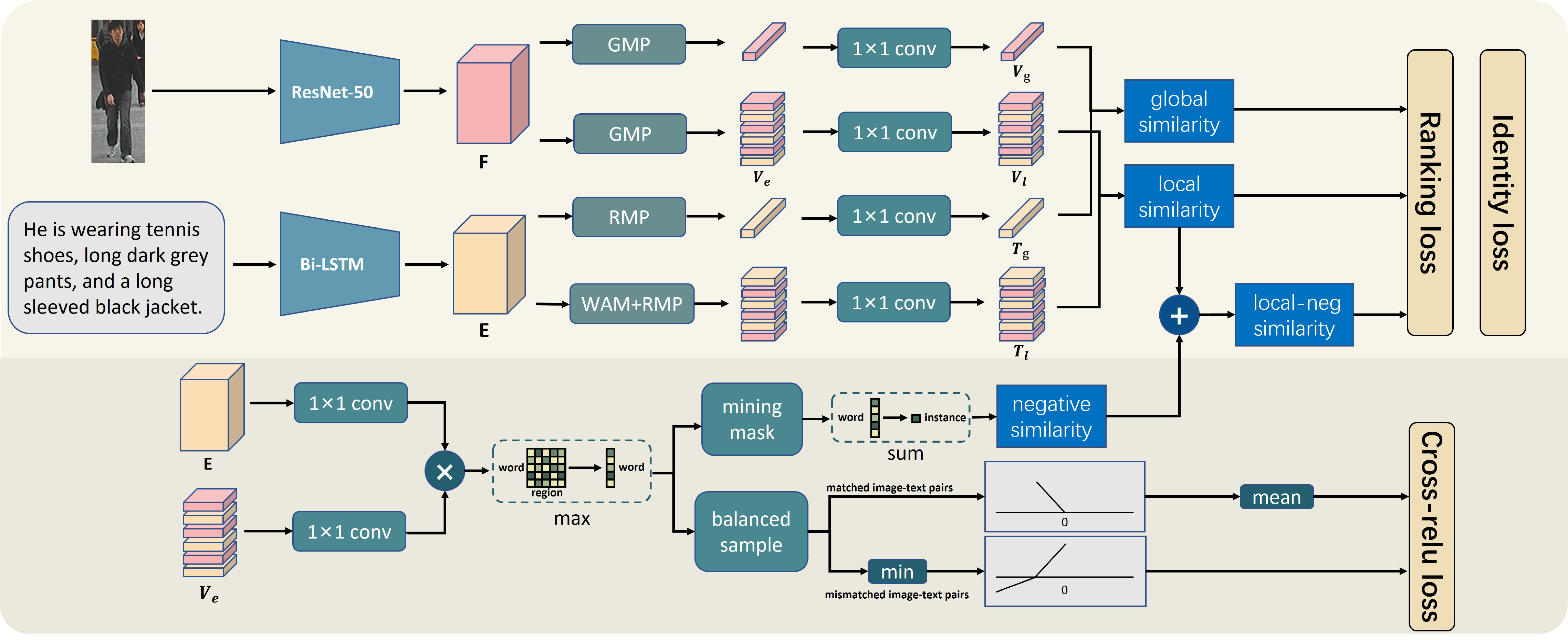}}
	\end{minipage}
	\caption{The overall framework of our proposed MFPE model, containing a global branch, a local branch, and an FPM branch to jointly infer the similarity of an image-text pair.}
	\label{method}
\end{figure*} We also use RMP and $K$ corresponding 1$\times$1 conv layers on modified representations to get final local textual representations.

When it comes to calculating similarity, we respectively concatenate the $K$ regions visual and textual representations $V_{l} $ and $T_{l} \in \mathbb{R} ^{K\times P} $ into the unified representations. Eventually, the global and local similarity of an image-text pair can be calculated as:
\begin{equation}
s_{g}=\frac{V_{g}^{T}T_{g}}{\left \| V_{g}  \right \|\left \| T_{g}  \right \|  },
	\hspace{1em}
s_{l}=\frac{V_{l}^{T}T_{l}}{\left \| V_{l}  \right \|\left \| T_{l}  \right \|  }.
\end{equation}
\subsection{FPM branch}
As discussed in the Introduction, the goal of the FPM branch is to highlight the effect of mismatched word-region pairs in an effective way. The first step is to mine the mismatched word-region pairs through the similarity scores. Concretely, we separately project local visual representations $V_{e} = \left \{v_{k} \right \}_{k=1}^{K}$ and textual representations $E = \left \{e_{i} \right \}_{i=1}^{n}$ into a common feature space via a 1$\times$	1 conv layer and then compute the semantic similarity scores as: 
\begin{equation}
s_{k,i}=\frac{\theta (v_{k})^{T}\phi(e_{i})}{\left \| \theta (v_{k})  \right \|\left \| \phi(e_{i})  \right \|  },k\in [1,K],i\in [1,n],
\end{equation}
where $\theta (v_{k})=W_{\theta }v_{k}$, $\phi(e_{i})=W_{\phi }e_{i}$. $W_{\theta },W_{\phi }\in \mathbb{R}^{M\times C}$. 

For text-based ReID, the number of regions is so limited that the region that mismatches any word is almost non-existent. Therefore, we mine mismatched word-region pairs by searching for words that mismatch any region. we perform a max-pooling operation on similarity scores because the maximum similarity between the word and all regions is low, indicating that the word mismatches any region.
\begin{equation}
s_{i}=\underset{k}{max}(s_{k,i}).
\end{equation}

The experimental analysis in section 4.3 demonstrates that a learnable decision boundary is unnecessary, so zero is considered as the decision boundary for matched and mismatched word-region pairs. In order to enhance the discriminative power of mismatched pairs, the {\bf mining mask} operation is employed to filter out negative similarities. Eventually, we use them to calculate the instance-level similarity and modify the local similarity:
\begin{equation}
s_{neg}=\sum_{i=1}^{n}Mask_{mining}(s_{i}),
	\hspace{1em}
s_{local-neg}=s_{l}+s_{neg},     
\end{equation}
where $Mask_{mining}(\cdot)$ denotes that when the input is positive, it is 0, and when the input is negative, it is unchanged. 

During inference, the overall similarity of an image-text pair is the sum of $s_{g} $, $s_{l}$, and $s_{local-neg}$.
\subsection{Optimization}
Instead of simply utilizing the similarity scores to distinguish whether the word-region pairs match, we propose a novel {\bf cross-relu loss} to increase the similarity gap between them. Specifically, We first adopt a {\bf balanced sample} strategy to obtain the same number of matched and mismatched image-text pairs in a mini-batch. Since there are no labels for word-region pairs, the coarse judgment of whether word-region pairs match is based on the following principles: 
1)	All word-region pairs of a matched image-text pair match.
2)	Mismatched image-text pair has at least one mismatched word-region pair.

For the matched image-text pairs, we adopt a relu with slope $m_{1}$ and bias $b_{1}$ to train their similarity scores tending to be positive:
\begin{equation}
L_{m}=\frac{1}{n}\sum_{i=1}^{n}max(-m_{1}s_{i}+b_{1},0),     
\end{equation}
where slope $m_{1}$ and bias $b_{1}$ are empirically set to 1 and 0.001 (The function is shown in Fig. \ref{method}) to ensure that the similarity scores keep away from the sensitive decision boundary.

For the mismatched image-text pairs, we first perform a min operation to obtain the most probable mismatched word-region pair. Similarly, we train it to be negative via a leaky relu with slope $m_{2}$ and bias $b_{2}$:
\begin{equation}
s_{min}=\underset{i}{min}(s_{i} ),     
	\hspace{1em}
L_{mm}=max(m_{2}s_{min}+b_{2},0),      
\end{equation}
where slope $m_{2}$ and bias $b_{2}$ are empirically set to 1 and 0.15. Note that adopting leaky relu instead of relu and setting $b_{2}$ large enough is aimed at maximizing the mining of mismatched word-region pairs at the cost of the accuracy of matched pairs.

To further optimize MFPE, the popular identity loss is employed in global and local branches.
\begin{equation}
L_{id}(x)=-log(softmax(W_{id}x)),   
\end{equation}
\begin{equation}
L_{id}=L_{id}(V_{g})+L_{id}(T_{g})+\lambda _{1}(L_{id}(V_{l})+L_{id}(T_{l})),
\end{equation}
where $W_{id}$ is a shared FC layer between the two modalities and $\lambda _{1}$ is set to 0.5.

Besides, the popular ranking loss is adopted to constraint the intra-class similarity score to be larger than the inter-class similarity with a margin $\alpha$:
\begin{equation}
	\begin{aligned}
L_{r}(s)&=max(\alpha-s(V_{+},T_{+})+s(V_{+},T_{-}),0)\\
&+max(\alpha-s(V_{+},T_{+})+s(V_{-},T_{+}),0), 
	\end{aligned}
\end{equation}
\begin{equation}
L_{r}=L_{r}(s_{g})+\lambda_{2}L_{r}(s_{l})+\lambda_{3}L_{r}(s_{local-neg}),
\end{equation}
where $(V_{+},T_{-})$ and $(V_{-},T_{+})$ denote a mismatched image-text pair while $(V_{+},T_{+})$ denotes a matched image-text pair. $S(\cdot,\cdot )$ stands for the similarity of a pair. Note that $\lambda_{2}$ and $\lambda_{3}$ are set to 0.5 and 0.25 to avoid $s_{local-neg}$ replacing $s_{l}$.

\section{EXPERIMENTS}
\subsection{Experiment Settings}
{\bf Dataset and Evaluation Metrics:} We conduct extensive experiments on CUHK-PEDES \cite{Li_2017_CVPR} dataset and adopt the popular Recall at K (R@K, K=1, 5, 10) to evaluate performance. Following the official evaluation protocol, the training set contains 34,054 images and 68,126 textual descriptions for 11003 persons. The validation and test sets include data for 1,000 persons, respectively. 
\\{\bf Implementation Detail:} All experiments are conducted on an NVIDIA Titan Xp GPU. We adopt Adam as the optimizer with the initial learning rate of 0.001. The mini-batch size and number of epochs are empirically set to 64 and 45. Following previous methods \cite{ding2021semantically}, $C$, $K$, $P$, $n$, $\alpha$, and $M$ are set to 2048, 6, 1024, 100, 0.2, and 256, respectively.
\subsection{Comparison Results}
We compare our proposed MFPE method with previous approaches on the CUHK-PEDES database in Table \ref{tab_com}. Comparison results show MFPE outperforms all other methods. For example, compared with the typical region-based model MIA, MFPE obtains a significant 10.72\% improvement on R@1. Moreover, compared with SSAN, MFPE still achieves nearly 2.45\% improvement in terms of R@1.
\subsection{Ablation Study}
In the following, we conduct extensive ablation studies on CUHK-PEDES to analyze the effectiveness of each branch in Table \ref{tab_abl1}. It is obvious that the local similarity modified by negative similarity achieves a significant improvement of 1.40\% on R@1 compared with the unmodified. The `global + local + FPM' also promotes the R@1 accuracy of the `global + local' by 2.70\%. The above improvement strongly demonstrates the effectiveness of our proposed framework for mining false positive examples.

Moreover, we conduct a series of ablation studies to analyze the effectiveness of each component of the FPM branch in Table \ref{tab_abl2}. 1) The baseline model refers to MFPE without the FPM branch and BERT in the training stage. 2) When removing the ranking loss on local-negative similarity, robustness and performance are severely degraded since the modified local similarity lacks a direct constraint. 3) Mining mask is a key component in the FPM branch to emphasize the effect of mismatched pairs, without which the performance drops by 1.34\% 4) Without balanced sampling, the performance is slightly decreased, due to the apparent propensity for mismatched data. But the cost of time increases by about 4 times. 5) Adding a learnable decision boundary to MFPE will obtain suboptimal performance. However, The boundary is always kept around the initial value of zero. Consequently, the learnable decision boundary is unnecessary.

\begin{table}[h]
	\vspace{-1.0em}
	\centering
	\caption{Performance Comparisons on CUHK-PEDES.}
		\setlength{\tabcolsep}{1.3mm}{
	\begin{tabular}{l|c|ccc}
		\toprule
		Methods& ref & R@1 & R@5 & R@10\\
		\midrule
		GNA-RNN \cite{zheng2020dual} & CVPR17 & 19.05 & - & 53.64 \\
		CMPM/C \cite{zhang2018deep} & ECCV18 & 49.37 & 71.69 & 79.27 \\
		MIA \cite{MIA} & TIP20 & 53.10 & 75.00  & 82.90 \\
		SCAN \cite{lee2018stacked} & ECCV18 & 55.86 & 75.97  & 83.69 \\
		SUM \cite{sum} & KBS22 & 59.22 & 80.35 & 87.60 \\
		DSSL \cite{zhu2021dssl} & MM21 & 59.98 & 80.41 & 87.56 \\
		MGEL \cite{MGEL} & IJCAI21 & 60.27 & 80.01 & 86.74\\
		SSAN \cite{ding2021semantically} & arXiv21 & 61.37 & 80.15 & 86.73 \\
		LapsCore \cite{lapscore} & ICCV21 & 63.40 & - & 87.80 \\
		\midrule
		MFPE w/o BERT(ours) & - & 61.92 & 80.80 & 87.25\\
		{\bf MFPE(ours)} & - & {\bf 63.82} & {\bf 82.63} & {\bf 88.66} \\
		\bottomrule
	\end{tabular}}
	\label{tab_com}
\end{table}
\begin{table}[h]
	\vspace{-1.0em}
	\centering
	\caption{Ablation study about branches on CUHK-PEDES.}
	\setlength{\tabcolsep}{1.5mm}{
	\begin{tabular}{cccc|ccc}
		\toprule
		Global & Local & FPM & bert & R@1 & R@5 & R@10 \\
		\midrule	
		$\checkmark$ & - & - & - & 54.68 & 75.42 & 82.73 \\
		- & $\checkmark$ & - & - & 57.57 & 77.06 & 84.86 \\
		$\checkmark$ & $\checkmark$ & - & - & 59.22 & 78.72 & 85.65 \\
		- & $\checkmark$ & $\checkmark$ & - & 58.97 & 78.44 & 85.49 \\
		$\checkmark$ & $\checkmark$ & $\checkmark$ & - & 61.92 & 80.80 & 87.25
		\\
		$\checkmark$ & $\checkmark$ & $\checkmark$ &  $\checkmark$& 63.82 & 82.63 & 88.66
		\\
		\bottomrule
	\end{tabular}}
	\label{tab_abl1}
\end{table}
\begin{table}[!h]
	\vspace{-1.0em}
	\centering
	\caption{Ablation study about the FPM branch design on CUHK-PEDES.}
	\begin{tabular}{l|ccc}
		\toprule
		Methods & R@1 & R@5 & R@10\\
		\midrule
		Baseline & 59.22 & 78.72 & 85.65 \\
		w/o local-negative ranking loss & 59.81 & 79.22  & 86.24 \\
		w/o mining mask & 60.58 & 80.77  & 87.17 \\
		w/o balanced sample & 61.22 & 80.30  & 87.28 \\
		w learnable decision boundary & 61.47 & 80.70 & 87.35 \\
		Full & 61.92 & 80.80 & 87.25 \\
		\bottomrule
	\end{tabular}
	\label{tab_abl2}
\end{table}
\begin{figure}[!h]
	\begin{minipage}[t]{1.0\linewidth}
		\centering
		\centerline{\includegraphics[width=8.5cm]{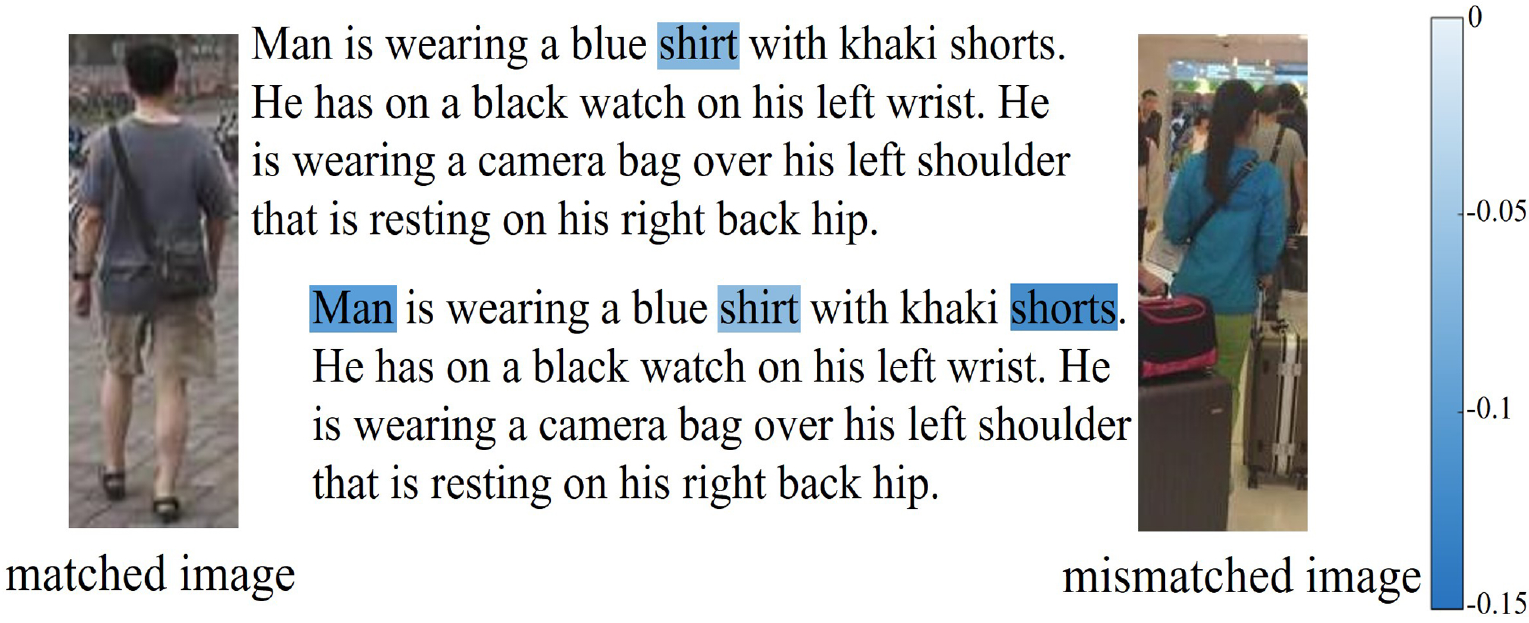}}
	\end{minipage}
	\caption{Visualization of the word-region negative similarity of a matched image-text pair and a mismatched image-text pair.}
	\label{visualize}
	\vspace{-1.0em}
\end{figure}
\subsection{Visualization}
We visualize the word-region negative similarity according to a given text query in Fig. 3. For the matched image, almost all words obtain non-negative similarity. As for the mismatched image, obviously mismatched words like `MAN' and `short' obtain a corresponding negative similarity. This case demonstrates the effectiveness of the FPM branch for mining mismatched pairs.
\section{CONCLUSION}
In this paper, we propose to mine false positive examples (MFPE) via a jointly optimized multi-branch architecture for text-based person ReID. Specifically, MFPE employs global and local branches to extract semantically aligned features and delicately designs an FPM branch to mine and emphasize the effect of mismatched word-region pairs. Moreover, We introduce a novel cross-relu loss to increase the gap of similarity scores between matched and mismatched word-region pairs. Finally, we conduct extensive experiments to demonstrate the effectiveness of MFPE.
%
\bibliographystyle{IEEEbib}
\bibliography{refs}

\begin{thebibliography}{10}

\bibitem{ding2021semantically}
Zefeng Ding, Changxing Ding, Zhiyin Shao, and Dacheng Tao,
\newblock ``Semantically self-aligned network for text-to-image part-aware
  person re-identification,''
\newblock {\em arXiv preprint arXiv:2107.12666}, 2021.

\bibitem{MIA}
Kai Niu, Yan Huang, Wanli Ouyang, and Liang Wang,
\newblock ``Improving description-based person re-identification by
  multi-granularity image-text alignments,''
\newblock {\em IEEE Transactions on Image Processing}, vol. 29, pp. 5542--5556,
  2020.

\bibitem{CMAMM}
Surbhi Aggarwal, Venkatesh~Babu Radhakrishnan, and Anirban Chakraborty,
\newblock ``Text-based person search via attribute-aided matching,''
\newblock in {\em Proceedings of the IEEE/CVF winter conference on applications
  of computer vision}, 2020, pp. 2617--2625.

\bibitem{zhu2021dssl}
Aichun Zhu, Zijie Wang, Yifeng Li, Xili Wan, Jing Jin, Tian Wang, Fangqiang Hu,
  and Gang Hua,
\newblock ``Dssl: Deep surroundings-person separation learning for text-based
  person retrieval,''
\newblock in {\em Proceedings of the 29th ACM International Conference on
  Multimedia}, 2021, pp. 209--217.

\bibitem{NAFS}
Chenyang Gao, Guanyu Cai, Xinyang Jiang, Feng Zheng, Jun Zhang, Yifei Gong, Pai
  Peng, Xiaowei Guo, and Xing Sun,
\newblock ``Contextual non-local alignment over full-scale representation for
  text-based person search,''
\newblock {\em arXiv preprint arXiv:2101.03036}, 2021.

\bibitem{HGAN}
Kecheng Zheng, Wu~Liu, Jiawei Liu, Zheng-Jun Zha, and Tao Mei,
\newblock ``Hierarchical gumbel attention network for text-based person
  search,''
\newblock in {\em Proceedings of the 28th ACM International Conference on
  Multimedia}, 2020, pp. 3441--3449.

\bibitem{TDE}
Kai Niu, Yan Huang, and Liang Wang,
\newblock ``Textual dependency embedding for person search by language,''
\newblock in {\em Proceedings of the 28th ACM International Conference on
  Multimedia}, 2020, pp. 4032--4040.

\bibitem{Li_2017_CVPR}
Shuang Li, Tong Xiao, Hongsheng Li, Bolei Zhou, Dayu Yue, and Xiaogang Wang,
\newblock ``Person search with natural language description,''
\newblock in {\em Proceedings of the IEEE Conference on Computer Vision and
  Pattern Recognition (CVPR)}, July 2017.

\bibitem{wang2020vitaa}
Zhe Wang, Zhiyuan Fang, Jun Wang, and Yezhou Yang,
\newblock ``Vitaa: Visual-textual attributes alignment in person search by
  natural language,''
\newblock in {\em European Conference on Computer Vision}. Springer, 2020, pp.
  402--420.

\bibitem{zhang2022negative}
Kun Zhang, Zhendong Mao, Quan Wang, and Yongdong Zhang,
\newblock ``Negative-aware attention framework for image-text matching,''
\newblock in {\em Proceedings of the IEEE/CVF Conference on Computer Vision and
  Pattern Recognition}, 2022, pp. 15661--15670.

\bibitem{he2016deep}
Kaiming He, Xiangyu Zhang, Shaoqing Ren, and Jian Sun,
\newblock ``Deep residual learning for image recognition,''
\newblock in {\em Proceedings of the IEEE conference on computer vision and
  pattern recognition}, 2016, pp. 770--778.

\bibitem{PCB}
Yifan Sun, Liang Zheng, Yi~Yang, Qi~Tian, and Shengjin Wang,
\newblock ``Beyond part models: Person retrieval with refined part pooling (and
  a strong convolutional baseline),''
\newblock in {\em Proceedings of the European conference on computer vision
  (ECCV)}, 2018, pp. 480--496.

\bibitem{lstm}
Sepp Hochreiter and J{\"u}rgen Schmidhuber,
\newblock ``Long short-term memory,''
\newblock {\em Neural computation}, vol. 9, no. 8, pp. 1735--1780, 1997.

\bibitem{bert}
Ashish Vaswani, Noam Shazeer, Niki Parmar, Jakob Uszkoreit, Llion Jones,
  Aidan~N Gomez, {\L}ukasz Kaiser, and Illia Polosukhin,
\newblock ``Attention is all you need,''
\newblock {\em Advances in neural information processing systems}, vol. 30,
  2017.

\bibitem{zheng2020dual}
Zhedong Zheng, Liang Zheng, Michael Garrett, Yi~Yang, Mingliang Xu, and Yi-Dong
  Shen,
\newblock ``Dual-path convolutional image-text embeddings with instance loss,''
\newblock {\em ACM Transactions on Multimedia Computing, Communications, and
  Applications (TOMM)}, vol. 16, no. 2, pp. 1--23, 2020.

\bibitem{zhang2018deep}
Ying Zhang and Huchuan Lu,
\newblock ``Deep cross-modal projection learning for image-text matching,''
\newblock in {\em Proceedings of the European conference on computer vision
  (ECCV)}, 2018, pp. 686--701.

\bibitem{lee2018stacked}
Kuang-Huei Lee, Xi~Chen, Gang Hua, Houdong Hu, and Xiaodong He,
\newblock ``Stacked cross attention for image-text matching,''
\newblock in {\em Proceedings of the European conference on computer vision
  (ECCV)}, 2018, pp. 201--216.

\bibitem{sum}
Zijie Wang, Aichun Zhu, Jingyi Xue, Daihong Jiang, Chao Liu, Yifeng Li, and
  Fangqiang Hu,
\newblock ``Sum: Serialized updating and matching for text-based person
  retrieval,''
\newblock {\em Knowledge-Based Systems}, vol. 248, pp. 108891, 2022.

\bibitem{MGEL}
Chengji Wang, Zhiming Luo, Yaojin Lin, and Shaozi Li,
\newblock ``Text-based person search via multi-granularity embedding
  learning.,''
\newblock in {\em IJCAI}, 2021, pp. 1068--1074.

\bibitem{lapscore}
Yushuang Wu, Zizheng Yan, Xiaoguang Han, Guanbin Li, Changqing Zou, and
  Shuguang Cui,
\newblock ``Lapscore: language-guided person search via color reasoning,''
\newblock in {\em Proceedings of the IEEE/CVF International Conference on
  Computer Vision}, 2021, pp. 1624--1633.

\end{thebibliography}

\end{document}